\documentclass[letterpaper,10pt,journal,twoside]{IEEEtran}  

\IEEEoverridecommandlockouts                              

\usepackage[T1]{fontenc}
\usepackage{lmodern}
\usepackage{booktabs}
\usepackage{multirow}
\usepackage{graphicx}
\usepackage{amsmath}
\usepackage{amssymb}
\usepackage{bm}
\usepackage{tabularx}
\usepackage{makecell}
\usepackage{pifont}
\usepackage{xcolor}
\newcommand{\cmark}{\textcolor{green!70!black}{\ding{51}}}
\newcommand{\xmark}{\textcolor{red!80!black}{\ding{55}}}
\title{\LARGE \bf
SpaceDex: Generalizable Dexterous Grasping in Tiered Workspaces
}

\author{Wensheng Wang$^{1,2}$, Chuanjun Guo$^{2}$, Wei Wei$^{2}$, Tong Wu$^{2}$, and Ning Tan$^{1}$%
\thanks{$^{1}$Sun Yat-sen University, Guangzhou, China.}%
\thanks{$^{2}$Stellarobot Company, Shenzhen, China.}%
}

\begin{document}

\maketitle
\thispagestyle{empty}
\pagestyle{empty}

\begin{abstract}

Generalizable grasping with high-degree-of-freedom (DoF) dexterous hands remains challenging in tiered workspaces, where occlusion, narrow clearances, and height-dependent constraints are substantially stronger than in open tabletop scenes. Most existing methods are evaluated in relatively unoccluded settings and typically do not explicitly model the distinct control requirements of arm navigation and hand articulation under spatial constraints.
We present SpaceDex, a hierarchical framework for dexterous manipulation in constrained 3D environments. At the high level, a Vision-Language Model (VLM) planner parses user intent, reasons about occlusion and height relations across multiple camera views, and generates target bounding boxes for zero-shot segmentation and mask tracking. This stage provides structured spatial guidance for downstream control instead of relying on single-view target selection.
At the low level, we introduce an arm-hand Feature Separation Network that decouples global trajectory control for the arm from geometry-aware grasp mode selection for the hand, reducing feature interference between reaching and grasping objectives. The controller further integrates multi-view perception, fingertip tactile sensing, and a small set of recovery demonstrations to improve robustness to partial observability and off-nominal contacts.
In 100 real-world trials involving over 30 unseen objects across four categories, SpaceDex achieves a 63.0\% success rate, compared with 39.0\% for a strong tabletop baseline. These results indicate that combining hierarchical spatial planning with arm-hand representation decoupling improves dexterous grasping performance in spatially constrained environments.

\end{abstract}

\section{INTRODUCTION}
Dexterous multi-fingered hands possess the kinematic versatility required to perform complex manipulation tasks similar to human capabilities. Recent advancements in data-driven methods, particularly those combining Vision-Language Models (VLMs) with imitation learning, have demonstrated impressive success in general dexterous grasping \cite{zhong2025dexgraspvla,brohan2023rt2,kim2024openvla}. These approaches effectively handle novel objects in cluttered scenarios, achieving high success rates by leveraging 2D foundational visual representations \cite{kirillov2023segment,ravi2024sam2segmentimages}. However, existing research predominantly operates under a significant simplification: the workspace is typically restricted to an open, flat tabletop where objects are visible from a top-down perspective and the robot arm moves freely without spatial confinement.

Real-world environments, such as warehouse shelves, domestic refrigerators, and tightly packed cabinets, introduce challenges that differ substantially from open tabletop scenes. In these spatially constrained environments, grasping is not only a reach-and-pick problem. First, target objects are often partially or fully occluded, which limits single-view perception. Second, shelf geometry imposes hard constraints: the robot must avoid collisions with walls and ceiling layers while respecting height-dependent dependencies among stacked objects. A practical system therefore needs both dexterous motor control and spatial reasoning to identify \emph{which} object blocks the target and \emph{how} to reach it.

Addressing these challenges requires a paradigm shift from 2D plane reasoning to 3D spatial awareness. Traditional motion planning algorithms can avoid obstacles but lack the semantic understanding to decompose complex clearing tasks. Conversely, end-to-end learning policies often struggle with the sim-to-real gap in sensing, especially when visual feedback is obstructed by the robot's own hand or environmental structures. Furthermore, without tactile feedback, a vision-based policy is prone to failure in confined spaces where precise contact is critical but visibility is compromised.

To bridge this gap, we present SpaceDex, a hierarchical Vision-Language-Action framework designed for robust dexterous grasping in multi-occluded and spatially constrained scenarios. Unlike prior works that treat the environment as a flat canvas, SpaceDex explicitly models the 3D structure and dependency relationships of the scene. Our framework comprises a High-Level Spatial Reasoner and a Low-Level Spatial-Tactile Controller.

At the high level, we leverage VLM reasoning to parse user instructions into actionable sub-tasks. The planner analyzes occlusion relationships and height disparities, generates a dependency graph, and prioritizes the removal of blocking objects before the final grasp. It then provides 2D bounding boxes (from the selected external camera view) and segmentation masks to ground the semantic plan in physical space.

At the low level, we introduce a controller for executing constrained grasp requests. To address visual information loss under occlusion, we combine a multi-view perception system (three third-person views plus one wrist view) with a feature disentanglement network that separates spatial geometry from visual texture. We also integrate fingertip tactile sensing to improve grasp stability when visual confirmation is limited. In addition, the controller includes a local recovery mechanism that can respond to slips or collisions without re-querying the high-level planner.

We evaluate SpaceDex in challenging setups with layered shelves and high-density clutter. On 100 real-world trials with over 30 unseen objects, our method reaches a 63.0\% success rate in constrained environments, compared with 39.0\% for a strong tabletop baseline. To our knowledge, SpaceDex is among the first frameworks to extend general dexterous grasping from open tabletop settings to constrained 3D shelving scenarios.

In summary, our key contributions are three-fold.
\begin{itemize}
\item We present SpaceDex, a hierarchical Vision-Language-Action framework that extends generalizable dexterous grasping from open tabletop scenes to spatially constrained 3D tiered workspaces.
\item We propose a Tier-Aware High-Level Planner that leverages multi-view VLM reasoning for occlusion handling, and a Low-Level Spatial-Tactile Controller featuring an Arm-Hand Feature Separation Network, tactile feedback, and recovery demonstrations.
\item We validate SpaceDex through extensive real-world experiments, achieving a 63.0\% success rate across 100 trials with over 30 unseen objects, outperforming a strong tabletop baseline by 24 percentage points. Ablation studies confirm the benefits of arm-hand feature separation, recovery data, and multi-camera perception.
\end{itemize}

\section{RELATED WORK}

\subsection{Dexterous Grasp Synthesis: Bridging Poses and Spatial Trajectories}
Traditional dexterous grasping studies primarily synthesize stable poses using geometric and physical metrics (e.g., force closure) \cite{prattichizzo2012manipulability,rodriguez2012caging}, often assuming known object geometry or simplified contact models. Data-driven approaches such as UniGrasp \cite{shao2020unigrasp} and DexMV \cite{qin2022dexmv} have extended grasp synthesis to multifingered hands by learning from object shape and human demonstrations. More recent policy-learning methods shift toward action-sequence generation and visuomotor control for open-world settings \cite{zhong2025dexgraspvla,chi2023diffusionpolicy,wen2025dexvla,zeng2022transporter,li2025softgrasp}. Nevertheless, many existing methods are still benchmarked in relatively open workspaces, where severe occlusions and tight geometric constraints are limited. In tiered shelves and cluttered cabinets, successful grasping requires trajectories that adapt online to height-dependent obstacles and narrow clearances, motivating tighter integration between spatial reasoning and dexterous control.

\subsection{VLM-Driven Task Decomposition and Spatial Reasoning}
Vision-Language Models (VLMs) are increasingly used for high-level robotic planning \cite{zhang2023llm}. Early work such as SayCan \cite{ahn2022saycan} and LLM-as-zero-shot-planners \cite{huang2022language} showed that large language models can decompose high-level instructions into actionable steps. Subsequent end-to-end vision-language-action models, including RT-2 \cite{brohan2023rt2}, PaLM-E \cite{driess2023palme}, and Octo \cite{ghosh2024octo}, scaled robotic control through web-scale pre-training and large multi-robot datasets \cite{oneill2024openx}. More recent open-source policies (OpenVLA \cite{kim2024openvla}, $\pi$0 \cite{black2024pi0}) further improved generalization. While these frameworks mainly use VLMs for object recognition or task sequencing, SpaceDex emphasizes spatial reasoning: by analyzing occlusion hierarchies and height relationships, the VLM provides spatial priors---such as execution priority and approach vectors---in addition to semantic labels.

\subsection{Generalizable Manipulation in Constrained Environments}
To improve generalization toward unseen objects, recent methods like DexGraspVLA utilize object masks and pre-trained features (e.g., DINOv2 \cite{oquab2023dinov2}, CLIP \cite{radford2021clip}) to focus attention on target objects. Large-scale imitation-learning datasets such as DROID \cite{khazatsky2024droid} and algorithms like ACT \cite{zhao2023act} have further improved sample efficiency in real-world manipulation. However, these approaches still struggle with environments governed by strict physical boundaries, such as shelves. SpaceDex addresses this gap through multi-modal perception and explicit arm-hand feature decoupling, enabling robust dexterous grasping in restricted spaces.

\section{METHOD}

\subsection{Problem Formulation}
Our goal is to develop a vision-based control policy for open-vocabulary dexterous grasping in spatially constrained 3D workspaces $\mathcal{W}$ (e.g., tiered shelving). We formulate this as a language-conditioned Partially Observable Markov Decision Process (POMDP).

Initially, a user provides a free-form language instruction $\mathcal{L}$. At each timestep $t$, the policy $\Pi$ receives an observation $\mathbf{o}_t = \{\mathbf{I}^{head}_t, \mathbf{I}^{wrist}_t, \boldsymbol{\tau}_t, \mathbf{s}_t\}$, where $\mathbf{I}^{head}_t = \{\mathbf{I}^{left}_t, \mathbf{I}^{front}_t, \mathbf{I}^{right}_t\}$ denotes three external RGBM views with $\mathbf{I}^{v}_t \in \mathbb{R}^{H \times W \times 4}$, $\mathbf{I}^{wrist}_t \in \mathbb{R}^{H \times W \times 3}$ is the wrist-view image, $\boldsymbol{\tau}_t \in \mathbb{R}^5$ denotes fingertip tactile feedback, and $\mathbf{s}_t = [\mathbf{s}^{arm}_t, \mathbf{s}^{hand}_t] \in \mathbb{R}^{22}$ represents the robot's proprioceptive state (7-DoF arm and 15-DoF hand).

The policy $\Pi$ generates an action trajectory $\boldsymbol{A}_t = \{\mathbf{a}_t, \dots, \mathbf{a}_{t+H_a-1}\}$ over horizon $H_a$, with each action decoupled as $\mathbf{a}_i = [\mathbf{a}^{arm}_i, \mathbf{a}^{hand}_i] \in \mathbb{R}^{22}$. Here $\mathbf{a}^{arm}_i$ dictates spatial positioning for collision avoidance, while $\mathbf{a}^{hand}_i$ controls fine-grained grasp actuation.

The framework operates hierarchically as $\Pi = \langle \pi_{high}, \pi_{low} \rangle$: a high-level VLM planner $\pi_{high}$ parses $\mathcal{L}$ to output target masks, while a low-level controller $\pi_{low}$ predicts the continuous trajectory $\boldsymbol{A}_t$ conditioned on observations and high-level guidance.

\begin{figure}[t]
\centering
\includegraphics[width=0.95\columnwidth]{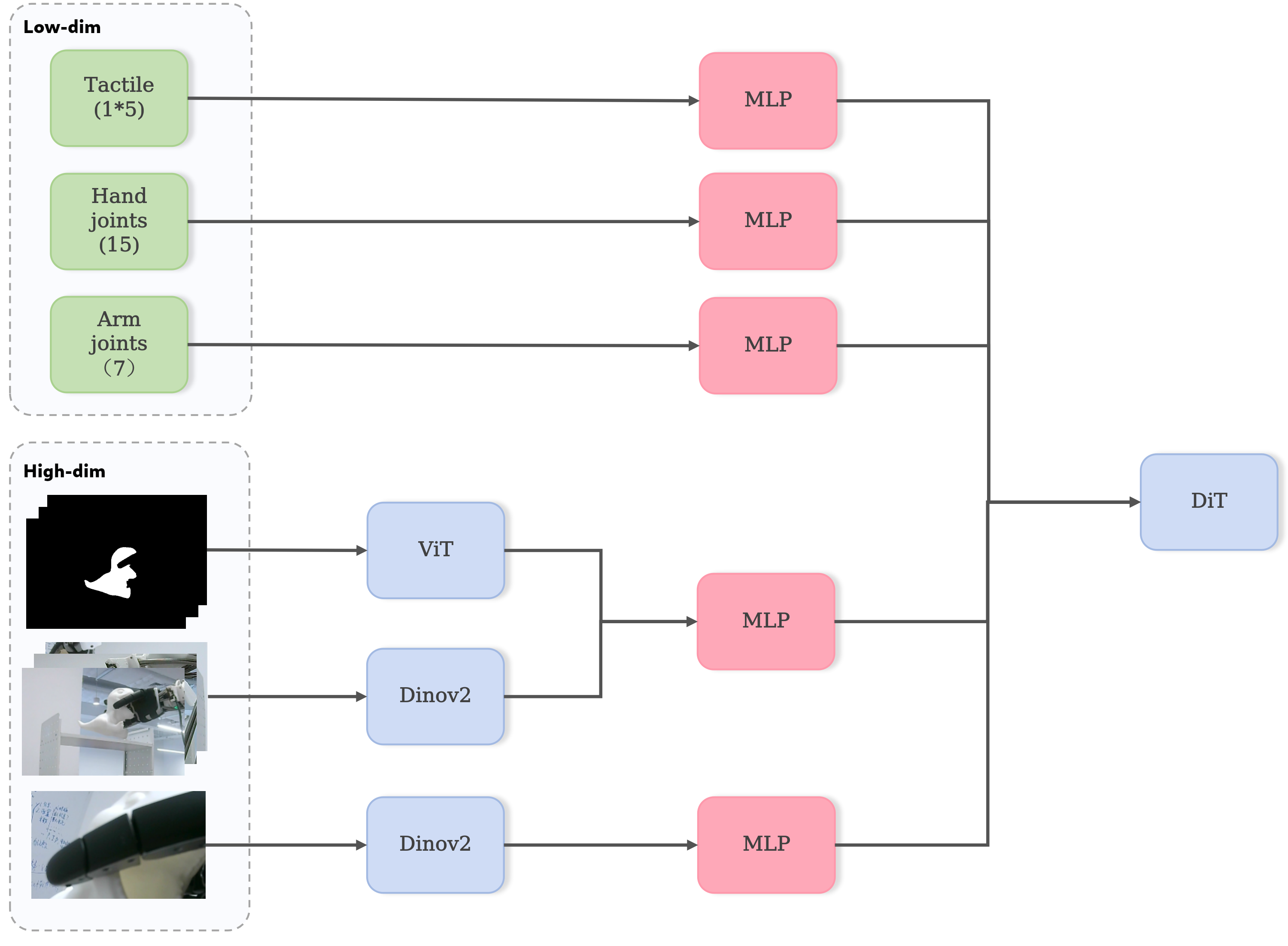}
\caption{Overview of the SpaceDex hierarchical framework. The high-level VLM planner performs tier-aware target selection and mask tracking across multiple camera views. The low-level diffusion controller generates collision-free trajectories and dexterous grasps conditioned on multi-modal observations.}
\label{fig:space_pipeline}
\end{figure}

\subsection{Data Collection and Recipe}
\label{sec:data_recipe}
High-quality demonstrations are important for robust imitation learning. To train SpaceDex, we designed a data collection protocol to improve generalization in real-world environments.

\textbf{Grasp-Prior-Guided Data Collection.} To avoid the prohibitive cost of full teleoperation for every object, we first generate geometric grasp priors using a distance-field network trained in our prior work. Each test object is scanned with a LiDAR-equipped tablet to obtain a watertight mesh, from which hundreds of physically feasible hand configurations are derived via inverse kinematics. These configurations serve as candidate grasps during subsequent human demonstration.

\textbf{Large-Scale Diverse Teleoperation.} We collected over 1,500 expert trajectories on the real robot platform across 3 different multi-tier shelves. The dataset encompasses more than 50 distinct everyday objects with significant variances in shape, texture, size, and deformability. 

\textbf{Error-Recovery Data Recipe.} A common bottleneck in Behavioral Cloning is covariate shift: once the policy deviates from the expert state distribution, errors compound and lead to failure. We include \textbf{approximately 5\% error-recovery demonstrations} in addition to standard successful trajectories. During collection, the operator introduces perturbations (e.g., dropping the object, misaligning the gripper, or colliding with the shelf edge) and then demonstrates corrective actions. As shown in Sec.~IV, this small recovery subset improves the policy's ability to recover from out-of-distribution states \cite{ross2011dagger,chi2023diffusionpolicy}.

\subsection{Tier-Aware High-Level Planning}
Standard VLM planners often fail in structured environments like tiered shelves due to depth ambiguity and self-occlusion. To address this, we introduce a \textbf{Tier-Aware Target Selection} strategy combined with \textbf{Parallel Multi-Camera Querying}.

Our perception system comprises three external fixed RGB cameras (DF100-720p, 2.8\,mm lens, 90$^\circ$ FoV) arranged at the left, front, and right of the shelf, plus one wrist-mounted camera for close-up manipulation views. Instead of naively prompting the VLM for a target, $\pi_{high}$ is instructed to reason about spatial dependencies (e.g., ``avoid objects that are blocked or surrounded''). Given the set of camera views $\mathcal{V} = \{v_1, v_2, v_3\}$, the planner independently processes each view in parallel using Qwen2.5-VL \cite{qwen2025vl72b} to predict bounding boxes $\mathbf{b}^{v_i}$ and confidence scores $c^{v_i}$. The optimal primary camera $v^*$ is selected via $v^* = \arg\max_{v_i} c^{v_i}$, subject to $c^{v^*} > 0.5$. This mechanism dynamically shifts the system's reliance to the least-occluded perspective, mitigating the ``blind spot'' issue inherent in dense shelving structures.

Once the primary view is selected, the Segment Anything Model (SAM) \cite{kirillov2023segment} generates an initial precise mask from the bounding box. To maintain temporal consistency across frames, we employ the Cutie video object segmentation network \cite{cheng2024cutie} to propagate the mask $\mathbf{M}_t$ through the episode. The resulting mask is concatenated with the head RGB stream to form an RGBM input $\mathbf{I}^{rgbm} \in \mathbb{R}^{518 \times 518 \times 4}$, grounding the semantic sub-task into the physical workspace for the low-level policy.

\begin{figure}[t]
\centering
\begin{minipage}[b]{0.31\columnwidth}
\centering
\includegraphics[width=\linewidth]{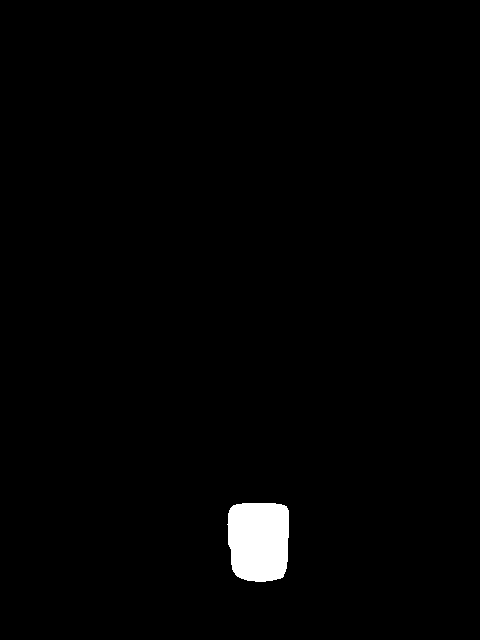}
\small (a) Left mask
\end{minipage}
\hfill
\begin{minipage}[b]{0.31\columnwidth}
\centering
\includegraphics[width=\linewidth]{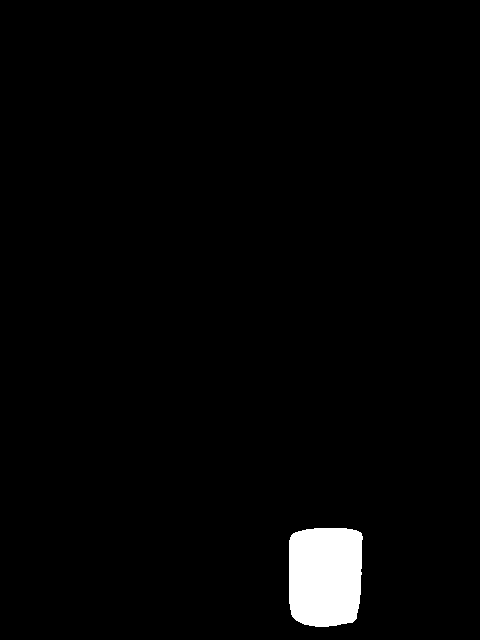}
\small (b) Center mask
\end{minipage}
\hfill
\begin{minipage}[b]{0.31\columnwidth}
\centering
\includegraphics[width=\linewidth]{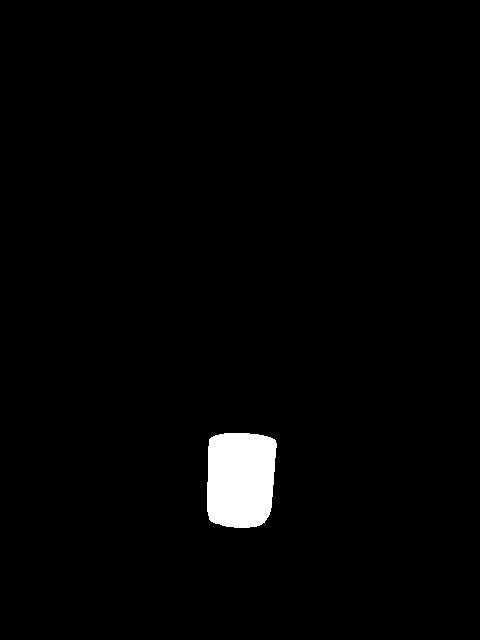}
\small (c) Right mask
\end{minipage}

\vspace{1mm}

\begin{minipage}[b]{0.31\columnwidth}
\centering
\includegraphics[width=\linewidth]{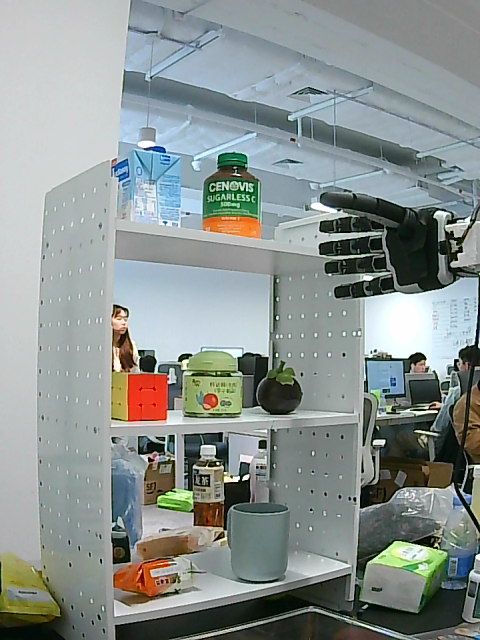}
\small (d) Left RGB
\end{minipage}
\hfill
\begin{minipage}[b]{0.31\columnwidth}
\centering
\includegraphics[width=\linewidth]{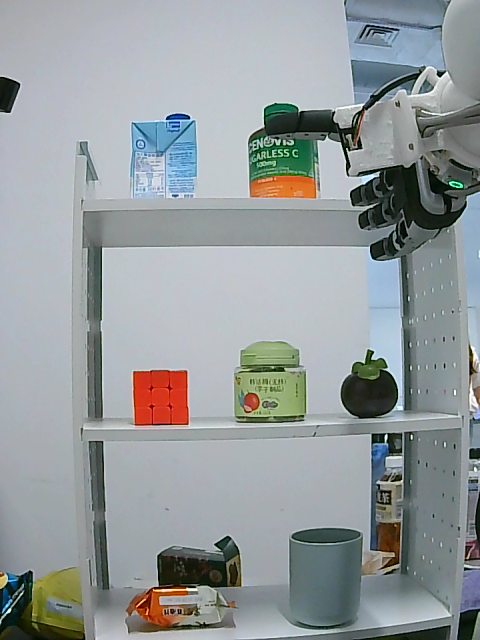}
\small (e) Center RGB
\end{minipage}
\hfill
\begin{minipage}[b]{0.31\columnwidth}
\centering
\includegraphics[width=\linewidth]{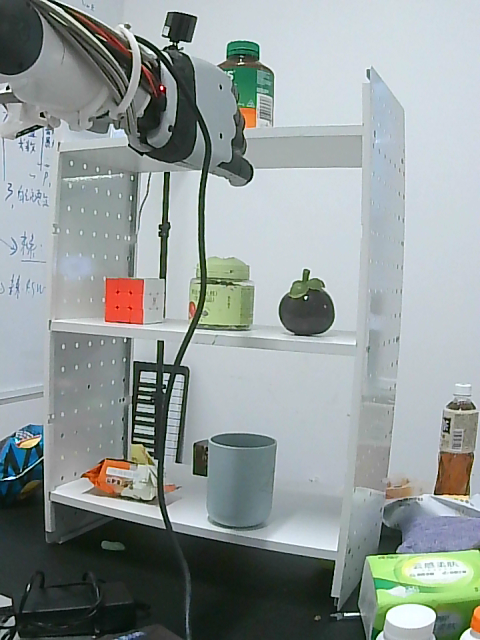}
\small (f) Right RGB
\end{minipage}
\caption{Multi-camera visual array and target segmentation masks. The top row shows the SAM-based masks and the bottom row shows the corresponding RGB images across the three camera views. The center view is selected as the primary camera because it exhibits the highest VLM confidence and the least occlusion.}
\label{fig:spacedex_cameras}
\vspace{-2mm}
\end{figure}

Fig.~\ref{fig:spacedex_cameras} illustrates the tier-aware planning process when the robot receives the instruction ``grasp the mug.'' The VLM queries all three external cameras in parallel; the center view yields the highest confidence because the mug is most clearly visible, whereas the side views suffer from perspective distortion and partial occlusion by the shelf frame.

\subsection{Spatially-Constrained Low-Level Control}
To robustly execute grasping trajectories in confined, multi-tiered workspaces, we formulate the low-level policy $\pi_{low}$ as a conditional diffusion process utilizing a Diffusion Transformer (DiT) backbone \cite{chi2023diffusionpolicy}. The model predicts a multi-step action chunk $\boldsymbol{A}_t = [\mathbf{a}_t, \dots, \mathbf{a}_{t+H_a-1}]$ with horizon $H_a=32$ at each timestep $t$ to ensure temporal smoothness and avoid collisions during reaching. 

\textbf{Mask Tracking and Feature Extraction:} 
The process begins by grounding the high-level semantic plan into the physical workspace. Given the target bounding box generated by $\pi_{high}$ at the initial timestep, we obtain a precise initial 2D mask of the target object and propagate it across frames as described above. 

Instead of naively stacking the mask as a fourth image channel, SpaceDex extracts visual and geometric cues independently using DINOv2 ViT-B/14 \cite{oquab2023dinov2} and a lightweight 4-layer mask encoder, and fuses them in latent token space to preserve high-fidelity semantics. For a $518\times518$ input, DINOv2 outputs $1\,369$ patch tokens of 768 dimensions. The selected primary external RGB image is mapped into patch tokens, while the tracked mask is encoded via the lightweight transformer. These are concatenated patch-wise to form a target-aware global feature $\mathbf{f}_{head}$. Simultaneously, the wrist camera image is processed by DINOv2 to yield fine-grained local features $\mathbf{f}_{wrist}$, augmented by an additional 3-layer transformer. These visual features, along with proprioception $\mathbf{s}_t$ and tactile feedback $\boldsymbol{\tau}_t$, are projected into a common latent space to form the multi-modal observation condition $\mathbf{C}_{obs}$.

The raw tactile signal originates from a $4\times3$ array on each fingertip (60D total). To reduce dimensionality and enhance generalization, we compute the L2 norm of each sensor's three axes and average across the four sensors per finger, yielding a 5D force-intensity vector $\boldsymbol{\tau}_t \in \mathbb{R}^5$.

\textbf{Diffusion-Based Action Generation:}
The control objective is to generate the action sequence $\boldsymbol{A}_0$ from Gaussian noise $\boldsymbol{A}_K$ through an iterative denoising process, conditioned on $\mathbf{C}_{obs}$:
\begin{equation}
    \boldsymbol{A}_{k-1} = \mathcal{D}_\theta(\boldsymbol{A}_k, k, \mathbf{C}_{obs})
\end{equation}
where $\mathcal{D}_\theta$ is the DiT-based denoising network, and $k$ represents the diffusion timestep. We adopt the DDIM scheduler \cite{song2021ddim} with 50 diffusion steps during training and 16 steps during inference for real-time responsiveness. Specifically, $\mathcal{D}_\theta$ first extracts a sequence of shared latent representations, which are subsequently decoupled to predict arm and hand actions (detailed in Sec.~III-E). During inference, we apply a receding horizon control strategy, executing only the first $H_a$ actions before re-planning.

\textbf{Wrist-Centric Cross-Attention:}
A common challenge in constrained shelf environments is self-occlusion, where the robot's arm frequently blocks the head camera's line of sight during deep reaching. To address this within the DiT architecture, we propose a \textit{Wrist-Centric Cross-Attention} mechanism. Instead of simple concatenation, we designate the wrist features as the Query ($Q$) to retrieve relevant spatial constraints from the global views (acting as Key $K$ and Value $V$):
\begin{equation}
    \mathbf{f}'_{wrist} = \mathbf{f}_{wrist} + \text{Attention}(Q=\mathbf{f}_{wrist}, K=\mathbf{f}_{ctx}, V=\mathbf{f}_{ctx})
\end{equation}
where $\mathbf{f}_{ctx}$ aggregates features from the other global camera views (e.g., the external-view feature $\mathbf{f}_{head}$, tactile $\boldsymbol{\tau}_t$, and proprioception $\mathbf{s}_t$). This cross-attention explicitly aligns the local manipulation perspective with global obstacle boundaries, ensuring that the policy remains spatially aware even when the primary external view is compromised.

To balance camera contributions, we apply importance weighting ($w_{wrist}=2.0$, $w_{head}=1.0$) and training-time camera dropout ($p_{drop}=0.5$ for external cameras), forcing the policy to remain robust to missing views.

\subsection{Arm-Hand Feature Separation Network}
A fundamental conflict exists in controlling a high-DoF system within confined spaces: the arm ($\mathbf{a}^{arm}$) requires global obstacle avoidance to navigate shelf layers, while the hand ($\mathbf{a}^{hand}$) requires fine-grained local geometry for grasping. A shared latent space often leads to feature interference, where dominant arm movements overshadow subtle hand articulations.

To resolve this, we introduce the Arm-Hand Feature Separation Network integrated into the DiT backbone. Let $\mathbf{X} \in \mathbb{R}^{H_a \times D}$ denote the sequence of latent tokens output by the shared transformer blocks ($D=768$). Instead of a unified projection, we bifurcate the representation learning into two specialized streams:
\begin{equation}
    \mathbf{h}_{arm} = \phi_{arm}(\mathbf{X}), \quad \mathbf{h}_{hand} = \phi_{hand}(\mathbf{X})
\end{equation}
Here, $\phi_{arm}$ and $\phi_{hand}$ are distinct two-layer encoding layers that project the shared tokens into 384-dimensional subspaces ($d_{arm}=d_{hand}=384$). The arm stream $\phi_{arm}$ focuses on spatial trajectory regression, adhering to shelf boundaries, whereas $\phi_{hand}$ focuses on object shape matching and contact optimization. 

To enforce that these streams learn distinct representations, we apply auxiliary supervision during training, requiring $\mathbf{h}_{arm}$ and $\mathbf{h}_{hand}$ to independently predict their respective noise components. The final action is synthesized by fusing these disentangled features with the original context:
\begin{equation}
    \hat{\boldsymbol{A}}_0 = \Psi([\mathbf{X}, \mathbf{h}_{arm}, \mathbf{h}_{hand}])
\end{equation}
where $\Psi$ is the final regression head mapping the $768+384+384=1\,536$-dimensional concatenated feature back to the 22-DoF action space. The auxiliary loss weight is set to $\lambda_{ah}=0.5$. This explicit separation allows SpaceDex to achieve precise collision-free reaching without sacrificing grasp stability.

\begin{figure*}[t]
\centering
\includegraphics[width=0.85\textwidth]{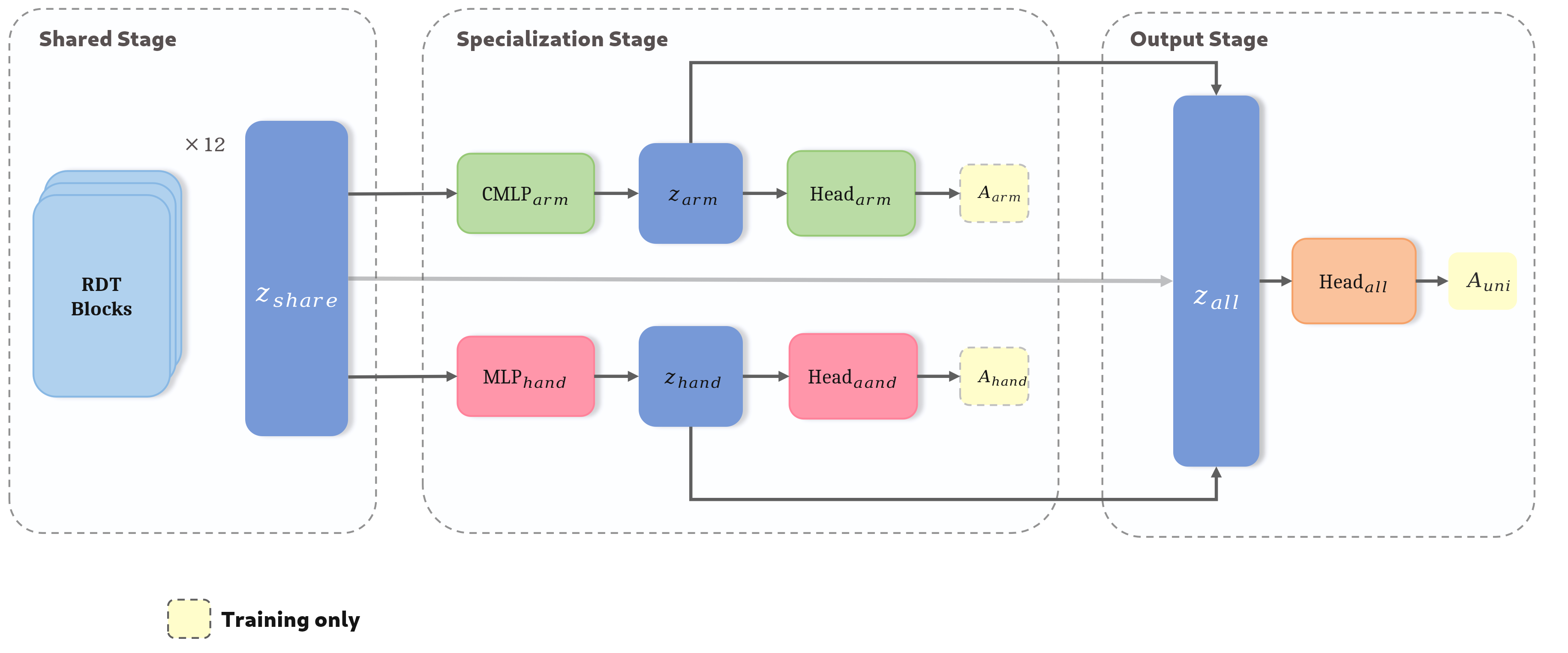}
\caption{Architecture of the Arm-Hand Feature Separation Network within the DiT policy. Shared latent tokens are split into arm and hand branches with task-specific auxiliary denoising supervision, then fused to regress 22-DoF actions for collision-aware reaching and stable grasping.}
\label{fig:space_seperate_pipeline}
\vspace{-3mm}
\end{figure*}

\section{EXPERIMENTS}

\subsection{Experimental Setup}
We evaluate SpaceDex on a real-world platform consisting of a 7-DoF RM75-B collaborative arm, a PantheonHand 15-DoF dexterous hand, and a three-layer shelf structure (bottom 30\,cm from ground, 25\,cm layer spacing, 40\,cm depth). Perception is provided by three external DF100-720p RGB cameras (left, front, right; 90$^\circ$ FoV) and one wrist-mounted camera. Fingertip 3D Hall-effect tactile sensors provide force feedback at 0.71\,mN resolution.

To evaluate grasping capability under varying physical constraints, we categorize test objects into four classes based on geometric and mechanical properties (Table~\ref{tab:object_classes}). The dataset comprises over 1,500 successful trajectories plus 5\% recovery demonstrations. Training uses the AdamW optimizer \cite{loshchilov2019adamw} with learning rate $1\times10^{-4}$, weight decay $1\times10^{-4}$, batch size 16, and 120 epochs with cosine annealing and 2,000-step warmup.

\begin{figure}[t]
\centering
\begin{minipage}[b]{0.47\columnwidth}
\centering
\includegraphics[width=\linewidth]{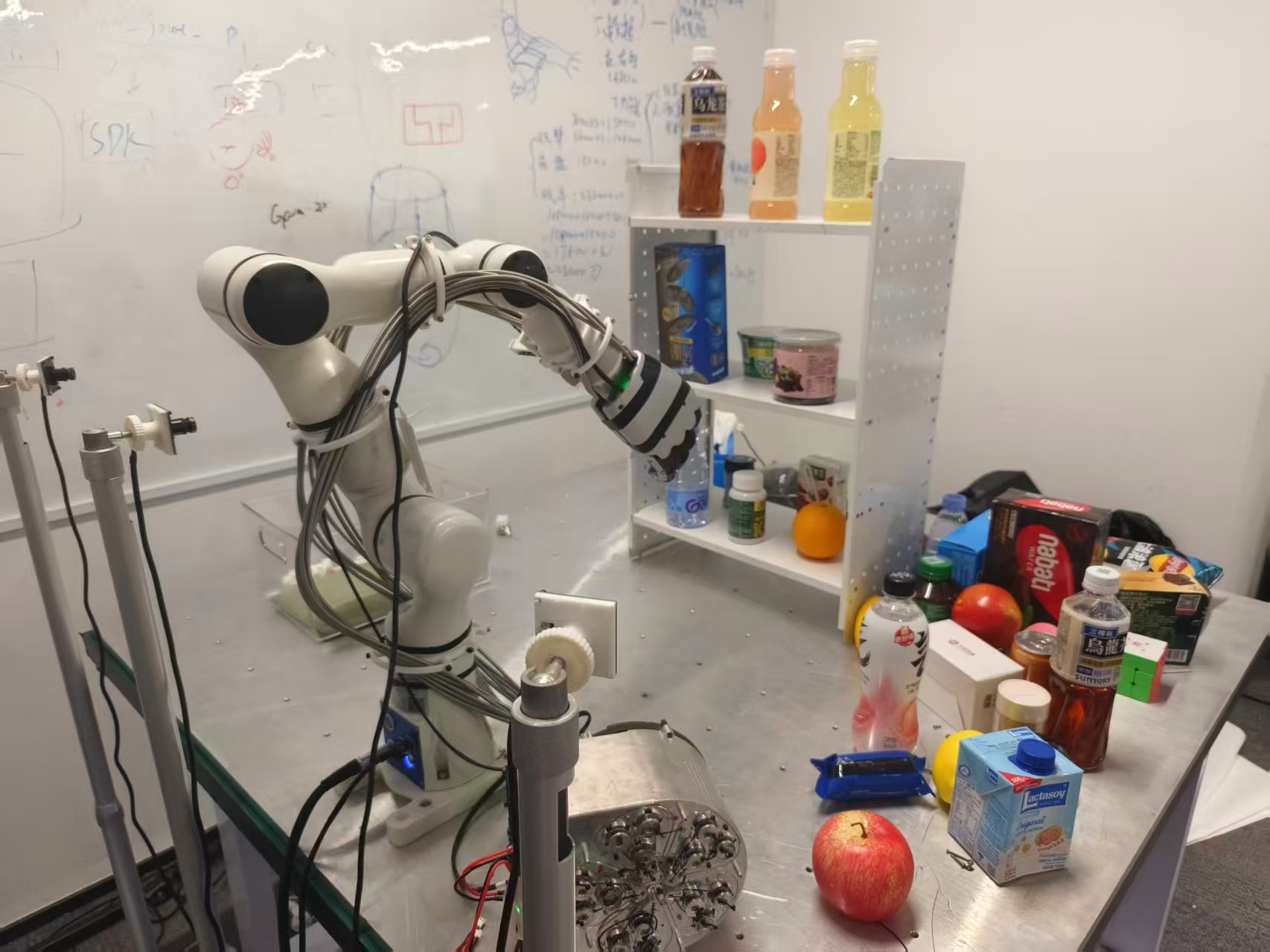}
\caption{Real-world experimental platform.}
\label{fig:setup}
\end{minipage}
\hfill
\begin{minipage}[b]{0.47\columnwidth}
\centering
\includegraphics[width=\linewidth]{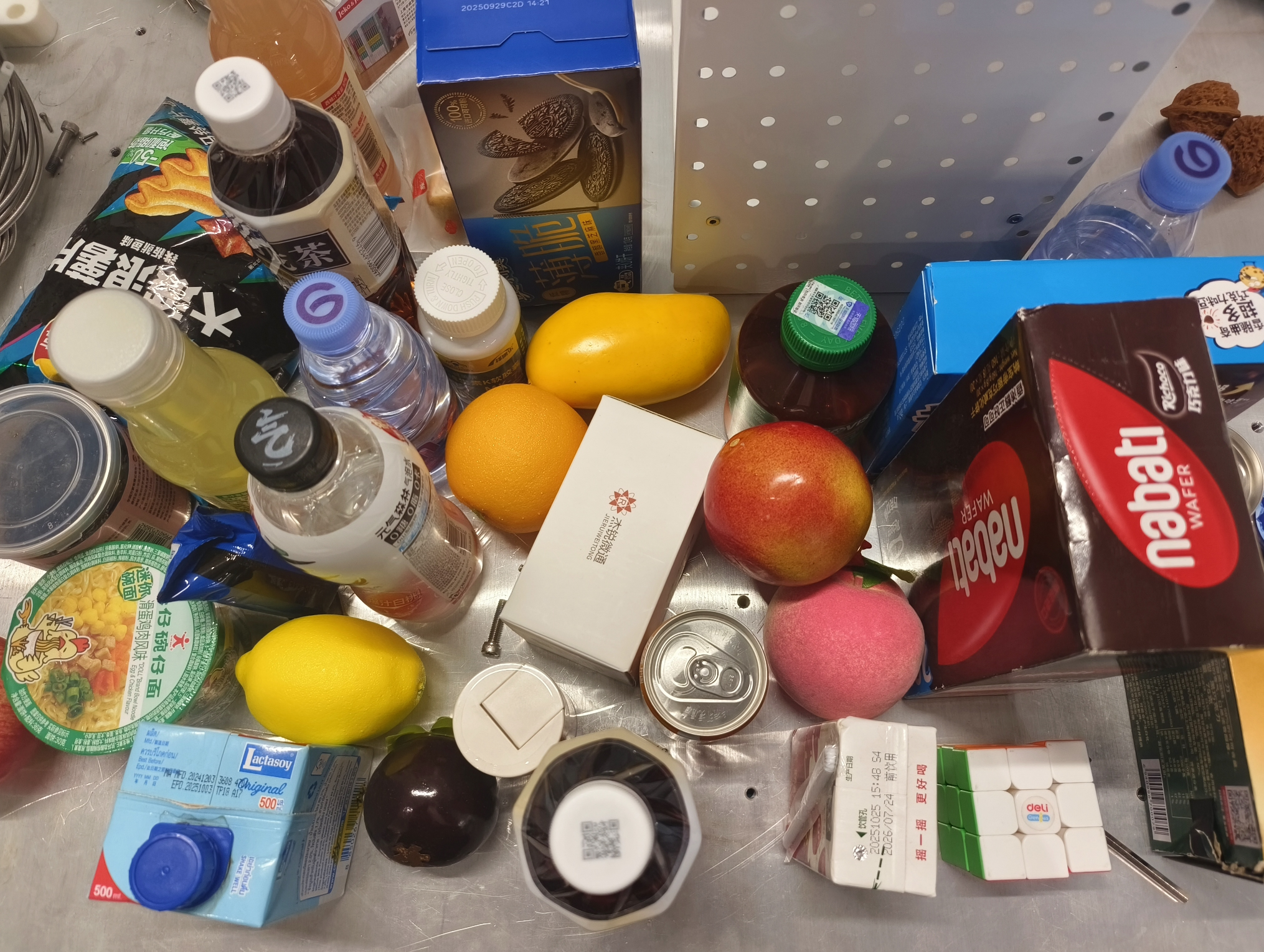}
\caption{Representative test objects.}
\label{fig:objects}
\end{minipage}
\end{figure}

\begin{table}[t]
\centering
\caption{Object categories and their grasping challenges.}
\label{tab:object_classes}
\small
\begin{tabularx}{\columnwidth}{l l X}
\toprule
\textbf{Category} & \textbf{Example} & \textbf{Grasping Challenge} \\
\midrule
Rigid Cuboids & Box, Cube & Flat faces facilitate contact planning. \\
Rigid Cylinders & Bottle, Can & Prone to rolling; requires lateral alignment. \\
Deformable & Bag, Carton & Easily deformed; low effective friction. \\
Near-spherical & Fruit & Slippery; contact points hard to localize. \\
\bottomrule
\end{tabularx}
\end{table}

\begin{table}[t]
\centering
\caption{Comparison of grasping success rate (SR \%) between DexGraspVLA and SpaceDex (Ours).}
\label{tab:grasping_performance}
\small
\setlength{\tabcolsep}{2pt}
\begin{tabular}{l c c c}
\toprule
\textbf{Object Category} & \textbf{Trials} & \textbf{DexGraspVLA} & \textbf{Ours} \\
\midrule
Rigid Cuboids & 25 & 32.0 & \textbf{48.0} \\
Rigid Cylinders & 25 & 56.0 & \textbf{80.0} \\
Near-spherical & 25 & 52.0 & \textbf{84.0} \\
Deformable & 25 & 16.0 & \textbf{40.0} \\
\midrule
\textbf{Overall Average} & 100 & 39.0 & \textbf{63.0 (+24.0)} \\
\bottomrule
\end{tabular}
\end{table}

\subsection{Main Results}

Prior to presenting quantitative results, we first compare SpaceDex with representative visuomotor policies from a methodological perspective. As summarized in Table~\ref{tab:method_comparison}, end-to-end VLA models such as RT-2 and OpenVLA rely on single-view perception and are evaluated mainly in open tabletop scenes, lacking tactile feedback and explicit arm-hand decoupling. Recent generalist policies including Octo, DexVLA, $\pi$0, and DexGraspVLA extend the policy class to multi-view inputs and dexterous hands, yet none of them incorporate fingertip tactile sensing or feature-level arm-hand separation, and they are not benchmarked in tiered or confined workspaces. In contrast, SpaceDex integrates all five key capabilities, making it the only framework that unifies hierarchical VLA planning, multi-view perception, tactile feedback, arm-hand decoupling, and dexterous control for constrained 3D environments.

\begin{table*}[t]
\centering
\caption{Methodological comparison between representative visuomotor policies and SpaceDex (Ours).}
\label{tab:method_comparison}
\small
\begin{tabular}{l c c c c c}
\toprule
\textbf{Method} & \textbf{VLM-based} & \textbf{Multi-view} & \textbf{Tactile} & \textbf{Arm-Hand} & \textbf{Dexterous} \\
 & \textbf{Planning} & \textbf{Perception} & \textbf{Feedback} & \textbf{Decoupling} & \textbf{Hand} \\
\midrule
RT-2~\cite{brohan2023rt2} & \cmark & \xmark & \xmark & \xmark & \xmark \\
OpenVLA~\cite{kim2024openvla} & \cmark & \xmark & \xmark & \xmark & \xmark \\
Octo~\cite{ghosh2024octo} & \xmark & \cmark & \xmark & \xmark & \cmark \\
DexVLA~\cite{wen2025dexvla} & \cmark & \cmark & \xmark & \xmark & \cmark \\
$\pi$0~\cite{black2024pi0} & \cmark & \cmark & \xmark & \xmark & \cmark \\
\textbf{DexGraspVLA}~\cite{zhong2025dexgraspvla} & \cmark & \cmark & \xmark & \xmark & \cmark \\
\midrule
\textbf{SpaceDex (Ours)} & \cmark & \cmark & \cmark & \cmark & \cmark \\
\bottomrule
\end{tabular}
\vspace{-2mm}
\end{table*}

Table~\ref{tab:grasping_performance} reports the real-world grasping success rates. Across 100 trials (25 per category), SpaceDex achieves an average success rate of 63.0\%, outperforming the DexGraspVLA baseline by 24 percentage points. The gains are most pronounced for near-spherical objects (84.0\% vs. 52.0\%) and deformable objects (40.0\% vs. 16.0\%), where tactile feedback and fine-grained hand control play critical roles.

\begin{figure}[t]
\centering
\begin{minipage}[b]{\columnwidth}
\centering
\includegraphics[width=0.95\linewidth]{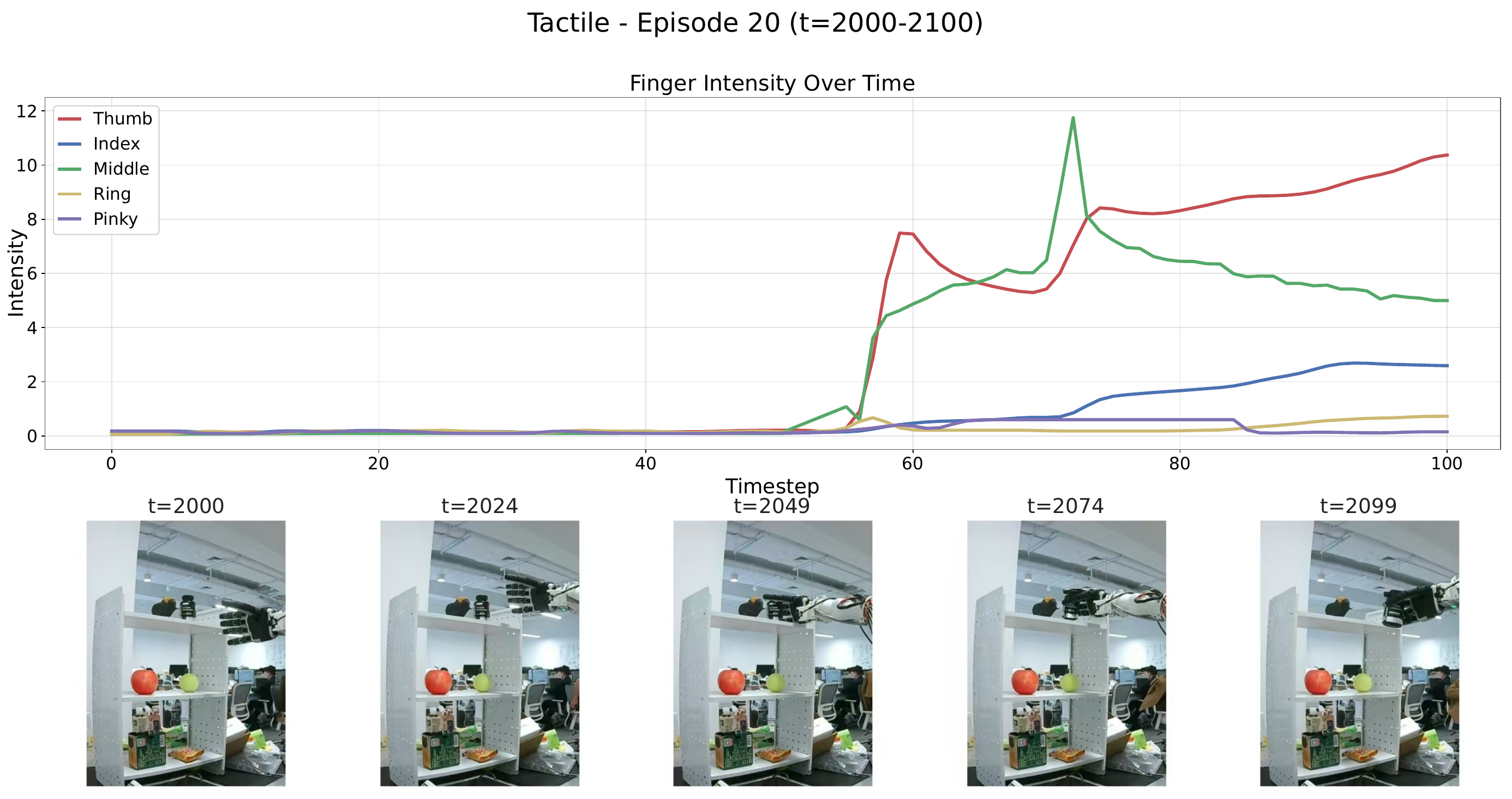}
\\[-0.4em]
{\small (a) Rigid cuboid grasp}
\end{minipage}

\vspace{2mm}
\begin{minipage}[b]{\columnwidth}
\centering
\includegraphics[width=0.95\linewidth]{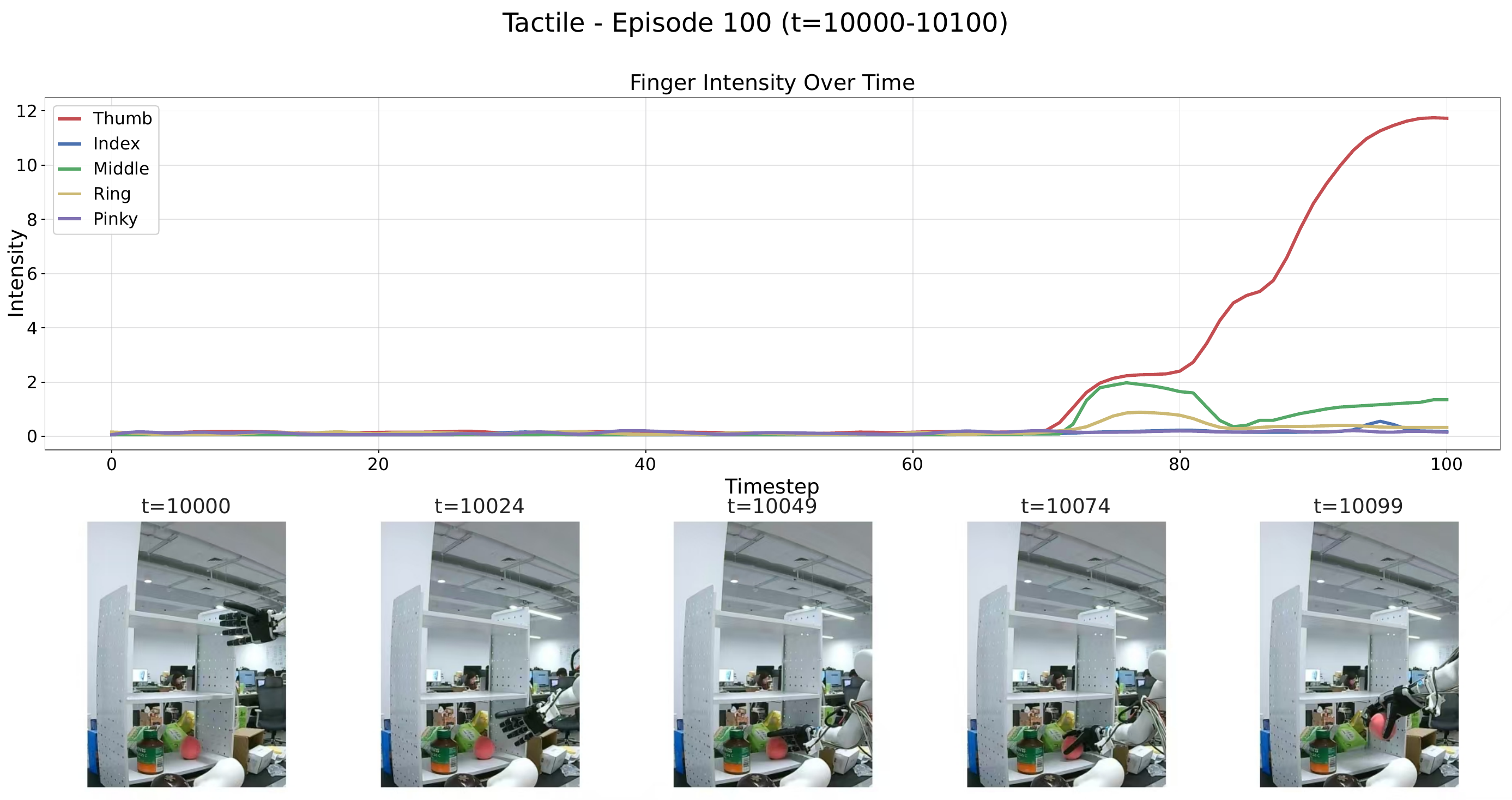}
\\[-0.4em]
{\small (b) Near-spherical object grasp}
\end{minipage}
\caption{Tactile force-intensity profiles during grasping of (a) a rigid cuboid and (b) a near-spherical object. The staggered contact pattern in (a) versus the synchronous rise in (b) demonstrates geometry-aware finger coordination.}
\label{fig:tactile_comparison}
\end{figure}

To further investigate the role of tactile sensing, Fig.~\ref{fig:tactile_comparison} plots the fingertip force-intensity profiles during typical grasps of a rigid cuboid and a near-spherical object. When grasping the cuboid, the thumb and middle finger establish contact first, followed by the index finger, producing a staggered rise in force that stabilizes the flat faces. In contrast, the spherical object induces nearly simultaneous contact across all fingers because the curved surface distributes contact points more uniformly. These divergent tactile signatures indicate that the policy leverages fingertip feedback to adapt closure patterns to object geometry.

\subsection{Ablation Studies}
\label{sec:ablation}
To evaluate the contribution of each module under spatial constraints and occlusion, we conduct ablation studies on the same 100 real-world trials. The quantitative results are summarized in Table~\ref{tab:ablation}.

\begin{table}[t]
\centering
\caption{Quantitative Results of Ablation Studies in Tiered Workspaces.}
\label{tab:ablation}
\small
\begin{tabularx}{\columnwidth}{X c c}
\toprule
\textbf{Model Variant} & \textbf{SR (\%)} $\uparrow$ & \textbf{Relative Drop (\%)} $\downarrow$ \\
\midrule
\textbf{SpaceDex (Full)} & \textbf{63.0} & --- \\
\midrule
w/o Arm-Hand Separation & 44.0 & 19.0 \\
w/o Recovery Data Recipe & 52.0 & 11.0 \\
w/o Multi-Camera (Single External) & 52.0 & 11.0 \\
\bottomrule
\end{tabularx}
\vspace{-2mm}
\end{table}

\textbf{1) Effect of Arm-Hand Feature Separation:} 
Removing the feature separation network drops the success rate from 63.0\% to 44.0\%. In tiered workspaces, the arm requires global spatial awareness to avoid shelf ceilings, while the hand requires localized geometric matching. Without decoupled encoding streams, \textit{feature interference} appears between spatial regression and fine hand articulation, resulting in trajectories that either graze shelf boundaries or fail to form force-closure grasps.

\textbf{2) Importance of the Error-Recovery Data Recipe:} 
Training exclusively on expert demonstrations degrades SR to 52.0\%. In narrow shelves, minor perception noise or physical perturbations can push the system into out-of-distribution states. Without recovery data, the policy more often exhibits erratic behavior after failure events. Adding 5\% corrective demonstrations improves closed-loop recovery.

\textbf{3) Impact of Multi-Camera Configuration:} 
Restricting the input to only a single external camera plus the wrist camera drops performance to 52.0\%. While the wrist camera provides excellent local textures for grasp execution, it lacks the global context required for safe path planning during the reaching phase. Our full multi-view setup ensures persistent 3D spatial awareness, enabling stable collision-free trajectories even when the target is deeply nested.

\subsection{Spatial Depth and Failure Analysis}
An implicit variable in tiered workspaces is the target depth. We observe a clear performance gradient across shelf layers: the bottom layer (least occluded) yields the highest success rates, while the top layer (deepest reach, maximum self-occlusion) presents the greatest challenge. Failures on the top layer are dominated by arm-shelf collisions rather than grasp quality, confirming that macro navigation is the bottleneck in deeply confined regions.

The most common failures involve \textit{deformable objects} (60\% failure rate within this category), where excessive finger closure causes the object to buckle or slip. Another frequent failure occurs during \textit{deep-layer reaching}: when the target is on the third shelf layer, the arm occasionally brushes against the upper shelf ceiling due to accumulated positioning error. Finally, \textit{tactile sensor saturation} on highly reflective or slippery surfaces occasionally prevents detecting initial contact, leading to delayed grasp closure. These cases point to future directions involving adaptive force control and finer depth estimation.

\section{DISCUSSION}

\subsection{Tiered vs. Open Workspaces}
The performance gap between SpaceDex and tabletop baselines reflects a clear difference between tiered and open workspaces. In open tabletop scenarios, the main challenge is object localization and grasp pose selection, which can often be handled with top-down cameras and simple reach-and-pick motions. In contrast, tiered shelves introduce occlusion, narrow clearances, and height-dependent obstacles that require 3D spatial reasoning. Our experiments show that tabletop policies such as DexGraspVLA perform less effectively in this setting (39.0\% vs. 63.0\%), likely because they lack multi-view redundancy for occlusion handling and do not disentangle arm navigation from hand grasping.

\subsection{Limitations and Broader Applicability}
While SpaceDex demonstrates strong performance in shelf-like scenarios, several limitations remain. First, our current validation is limited to static three-layer shelving; scalability to deeper, more irregularly nested structures remains to be explored. Second, the hierarchical design incurs non-negligible computation cost: VLM querying and DiT inference require GPU acceleration, which may limit deployment on resource-constrained edge robots. Finally, although our 15-DoF hand provides dexterity, extremely small or transparent objects still pose challenges for both vision and tactile sensing.

Despite these limitations, the core principles of SpaceDex---tier-aware semantic planning, multi-view perception, and arm-hand feature separation---transfer naturally to other constrained environments such as warehouse pick-and-place, domestic cabinet organization, and retail restocking. By decoupling high-level spatial reasoning from low-level dexterous control, our framework provides a general template for extending visuomotor policies from flat, open workspaces to complex, structured 3D environments.

\section{CONCLUSION}
We present SpaceDex, a dexterous grasping VLA framework for tiered workspaces. Building on recent advances in VLMs and diffusion policies, SpaceDex combines tier-aware planning with spatially constrained control and explicit arm-hand feature separation. In 100 real-world trials across four object categories, SpaceDex achieves a 63.0\% success rate, outperforming DexGraspVLA by 24 percentage points. Ablation studies show that the feature separation network, the recovery data recipe, and the multi-camera perception system each contribute to performance in confined spaces. These findings support the design of more generalizable manipulation systems for structured 3D environments.

\bibliographystyle{IEEEtran}
\bibliography{ref}

\end{document}